\documentclass[runningheads]{llncs}
\usepackage[T1]{fontenc}
\usepackage{graphicx}
\usepackage{subfigure}
\usepackage[ruled,vlined,linesnumbered]{algorithm2e}
\usepackage{multirow}
\usepackage{amsmath} 
\usepackage{setspace}
\usepackage{booktabs}

\usepackage{hyperref}
\hypersetup{hidelinks,
	colorlinks=true,
	allcolors=black,
	pdfstartview=Fit,
	breaklinks=true}
\usepackage{floatrow}

\usepackage{bm}
\usepackage{amsfonts} 
\usepackage{stmaryrd} 

\usepackage{mathrsfs} 
\usepackage{todonotes}
\usepackage{comment}
 
\usepackage{color}
\definecolor{Orange}{rgb}{1,0.5,0}
\definecolor{Red}{rgb}{1,0,0}
\definecolor{Blue}{rgb}{0,0,1}

\begin{document}
\title{Multi-Label Adaptive Batch Selection by Highlighting Hard and Imbalanced Samples}



\author{Ao Zhou\inst{1} \and
Bin Liu\inst{1}\thanks{Corresponding author.} \and
Jin Wang\inst{1} \and
Grigorios Tsoumakas\inst{2}}

\authorrunning{Ao Zhou et al.}

\institute{Key Laboratory of Data Engineering and Visual Computing,\\
Chongqing University of Posts and Telecommunications, China\\
\email{zacqupt@gmail.com, \{binliu, wangjin\}@cqupt.edu.cn}
\and
School of Informatics, Aristotle University of Thessaloniki, Greece\\
\email{greg@csd.auth.gr}}

\maketitle

\begin{abstract}
Deep neural network models have demonstrated their effectiveness in classifying multi-label data from various domains. Typically, they employ a training mode that combines mini-batches with optimizers, where each sample is randomly selected with equal probability when constructing mini-batches. However, the intrinsic class imbalance in multi-label data may bias the model towards majority labels, since samples relevant to minority labels may be underrepresented in each mini-batch. Meanwhile, during the training process, we observe that instances associated with minority labels tend to induce greater losses. Existing heuristic batch selection methods, such as priority selection of samples with high contribution to the objective function, i.e., samples with high loss, have been proven to accelerate convergence while reducing the loss and test error in single-label data. However, batch selection methods have not yet been applied and validated in multi-label data. In this study, we introduce a simple yet effective adaptive batch selection algorithm tailored to multi-label deep learning models. It adaptively selects each batch by prioritizing hard samples related to minority labels. A variant of our method also takes informative label correlations into consideration. Comprehensive experiments combining five multi-label deep learning models on thirteen benchmark datasets show that our method converges faster and performs better than random batch selection.

\end{abstract}

\section{Introduction}
Multi-label classification (MLC), which is concerned with assigning multiple labels to each instance at once, arises frequently as a task across various domains, including text categorization~\cite{text1}, bioinformatics~\cite{bio}, and automatic image labeling~\cite{image2}. For instance, a news article could span categories like {\em Technology}, {\em Economy}, and {\em Health}.

Deep learning has recently proven successful in addressing multi-label classification challenges \cite{C2AE,MPVAE,PACA,HOTVAE,DELA,CLIF}. 
By forming appropriate latent embedding spaces, deep neural networks manage to unravel the complex dependencies between features and labels in multi-label data~\cite{C2AE,MPVAE}. 
Moreover, deep learning models can successfully dissect and analyze label correlations \cite{CLIF,HOTVAE}. In addition, their inherent strength in representation learning allows them to naturally model label-specific features \cite{PACA}.

Recently, the issue of class imbalance in multi-label data has garnered growing interest \cite{mlcimbreview}. Due to the inherent nature of multi-label data, label frequencies (the number of instances relevant to each label) are significantly diverse. The high-frequent \textit{majority labels} typically dominate the training process of multi-label classifiers, leading to an undue ignorance of less-frequent \textit{minority labels}. Class imbalance impacts the training of deep learning models too. 
Conventional training of multi-label deep neural networks relies on an optimizer with randomly sampled mini-batches, i.e., every sample has an equal chance to be selected in the batch~\cite{long-tail}. 
Due to label rarity and batch size, instances associated with minority labels seldom appear in mini-batches over epochs.
In each epoch, the features of instances associated with \textit{minority labels} are exposed within fewer batches, leading to deep learning models inadequately learning more difficult minority labels.



\begin{figure}[!h]
    \centering
    \subfigure[30-th epoch]{\includegraphics[width=0.48\textwidth]{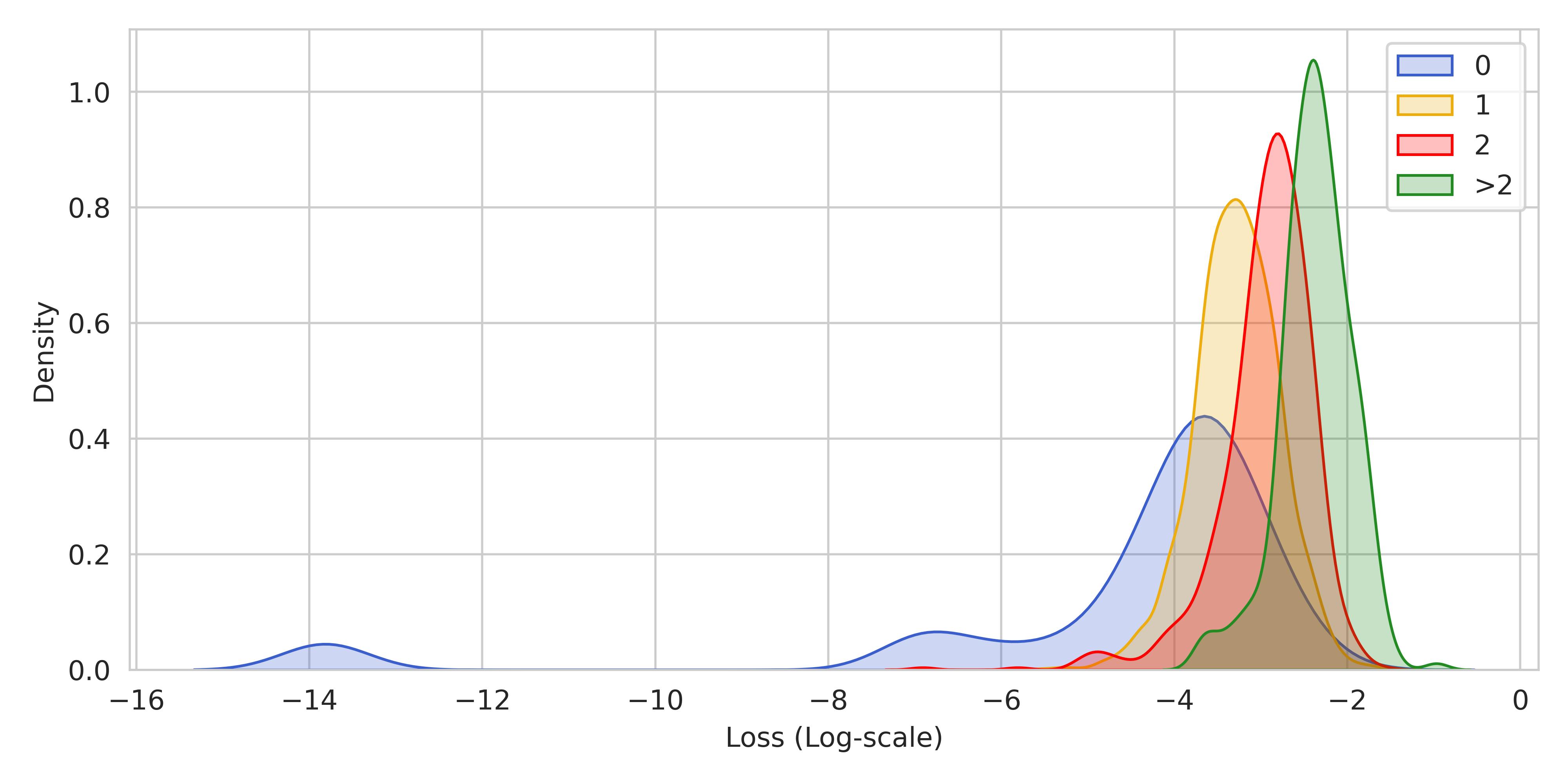}
    } 
    \subfigure[70-th epoch]
    {\includegraphics[width=0.48\textwidth]{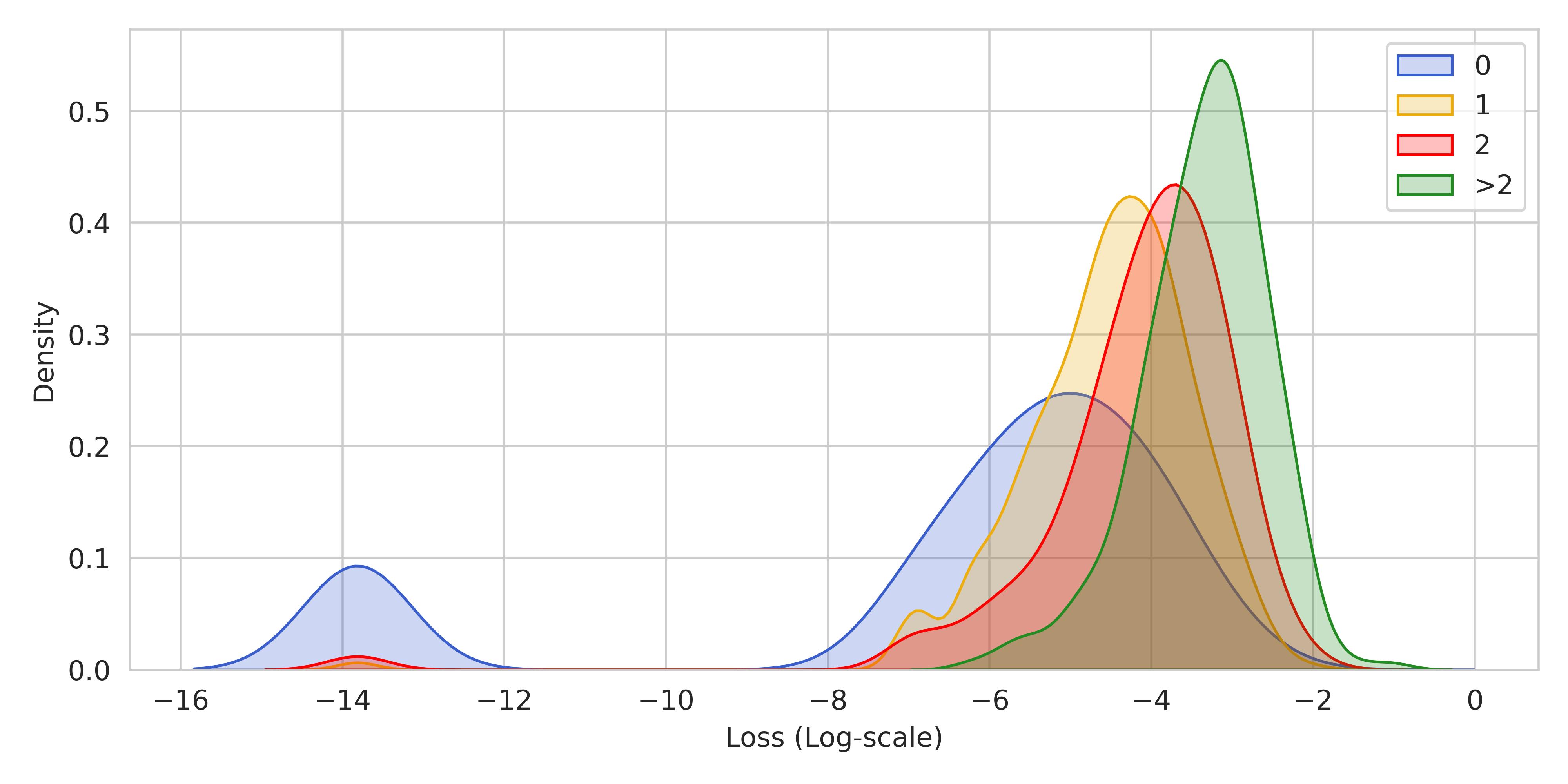}
    } 
    \caption{The log-scaled loss density distributions of the DELA model trained on the bibtex dataset at the 30-th and 70-th epochs, where blue, yellow, red and green curves denote the loss distribution of instances associated with zero, one, two and more than two \textit{minority labels}, respectively.}
    \label{fig:lossdistribution}
\end{figure}

Figure \ref{fig:lossdistribution} shows the loss density distributions of the DELA \cite{DELA} model on the bibtex dataset during the 30-th and 70-th epoch across samples associated with different numbers of \textit{minority labels}. 
Obviously, instances associated with more \textit{minority labels} tend to suffer higher losses throughout the entire training process. 
The connection between the higher loss and minority labels offers valuable insight: similar to higher loss samples, instances associated with minority labels are more crucial for training models.  
Samples with higher loss, also called hard samples, pose a challenge for the deep learning models. It has been demonstrated that emphasis on hard samples in batch selection can significantly improve the generalization ability \cite{OHEM} and speed up the convergence of deep learning models in single-label data scenario \cite{Onlinebatchselection,batch_quality1,batch_quality2}. 
However, choosing proper multi-label samples for each batch has not been investigated yet.


In this paper, we introduce an adaptive batch selection algorithm tailored for multi-label scenarios, designed to boost the training efficiency and performance of deep learning models. 
Our approach utilizes the widely adopted multi-label binary cross entropy loss to assess the difficulty of each sample, and imposes more attention on samples suffering global and local imbalance. The goal is to derive an adaptive loss that calculates the selection probability of each sample without adding to the time overhead. 
In addition, by adopting quantization index-based probability assignment, batch selection becomes insensitive to the small loss change and results in a smoother overall probability distribution. In addition, a variant of our method leverages a chain-based selection strategy to exploit label correlations explicitly. 
Integrating the adaptive batch selection method with five multi-label deep learning models achieves significantly improved performance and faster convergence than the default batch selection strategy.

\section{Related Work \label{related work}}
\subsection{Multi-Label Classification}
Multi-label classification, leverages techniques that harness label correlations to streamline the learning process and navigate complexity through effective use of label relationships \cite{mlcreview1,mlcreview2}.
Broadly, these methodologies can be categorized based on the complexity of label correlations they consider: first-order (e.g., BR, MLkNN \cite{BR,MLKNN}), second-order (e.g., CLR \cite{CLR}), and high-order correlations (e.g., CC, RAkEL \cite{cc,RAkEL}). 

Recently, deep learning has become a successful technique to solve the multi-label classification problem
BP-MLL \cite{BP-MLL} is the first neural network architecture for multi-label learning, which develops a pairwise loss function to explore label dependencies. 
Deep embedding-based methods have demonstrated effectiveness by using deep neural networks to harmonize the latent spaces of features and labels. For instance,
C2AE \cite{C2AE} embeds both feature and label data into deep latent space, employing a label-correlation sensitive loss function for end-to-end label output prediction.
MPVAE \cite{MPVAE} aligns probabilistic embedding spaces for labels and features, utilizing a decoder to model the joint distribution of output targets based on a multivariate probit model, emphasizing label correlation capture.
Several works focus on exploiting deep neural networks to capture label correlations. For instance,
PACA \cite{PACA} creates a latent metric space regulated by label correlation, learning prototypes and metrics for each label, and employs normalizing flows for capturing class label characteristics. 
HOT-VAE \cite{HOTVAE} is an attention-based framework for multi-label classification, which innovatively learns high-order label correlations adaptive to feature changes. 
Additionally, due to the powerful representation learning capability of deep neural networks, it is quite natural to consider the problem of label-specific features in the deep learning scenario. For instance,
CLIF \cite{CLIF} merges label semantics learning with the extraction of label-specific features, incorporating a graph autoencoder for semantic relations and a module for disentangling label-specific features.
DELA \cite{DELA} implements a perturbation-based technique for label-specific feature stability within a probabilistically relaxed expected risk minimization framework.

The imbalanced approaches proposed for MLC can be divided into three categories: sampling methods, classifier adaptation, and ensemble approaches \cite{mlcimbreview}. 
Sampling methods aim to balance the label distribution by either oversampling \cite{MLSMOTE,MLSOL,MLBOTE} minority labels or undersampling \cite{MLeNN,MLTL} majority ones, thus aiming to produce the more balanced versions of the training set before training. 
Classifier adaptation \cite{SOSHF} techniques modify existing algorithms to make them more sensitive to label imbalance, often by adjusting decision thresholds or incorporating imbalance-aware loss functions. 
Ensemble approaches \cite{ECCRU3} combine multiple models or algorithms to leverage their collective strength, often incorporating mechanisms to specifically address label imbalance, such as weighted voting or selective ensemble training focused on underrepresented labels.

In the field of multi-label learning, the existing batch selection algorithms are only applicable to active multi-label learning tasks \cite{active2,active1}, which are different from the scope of our multi-label classification. Active multi-label learning aims to progressively select unlabeled samples with the maximum information for manual annotation. 

\subsection{Hard Sample and Batch Selection in Single Label Data}

Hard samples are crucial for deep learning as they drive the model to refine its decision boundaries and improve generalization, making the learning process more effective and robust \cite{OHEM,hardsample,hardsample2}. For example, OHEM (Online Hard Example Mining) \cite{OHEM} focuses on training with hard samples, identified by their high losses, and exclusively using these instances to update the model's gradients.

Recent studies have pointed out that the performance of deep neural networks depends heavily on how well the mini-batch samples are selected. Meanwhile, it has been verified that neural networks converge faster with the help of intelligent batch selection strategies \cite{batch_quality1,batch_quality2,RecencyBias}. 
Techniques that select mini-batches based on sample difficulty have enhanced network precision and hastened training convergence. For instance, Online Batch Selection \cite{Onlinebatchselection} boosts training efficiency by ranking samples according to their recent loss values and adjusting their selection probability with an exponential decay based on rank. This approach strategically prioritizes higher-loss samples for upcoming mini-batches. 
Further, Recency Bias \cite{RecencyBias} targets samples with fluctuating predictions, using recent history to gauge uncertainty and increase their sampling probability for upcoming batches. 
Ada-Boundary \cite{Ada-boundary} prioritizes samples near the decision boundary where the model is uncertain, accelerating convergence by focusing on these critical samples. 
In multi-label settings, the complexity of measuring uncertainty for each label and aggregating these measures significantly increases computational demands. Additionally, the inherent sparsity and inter-label correlations in multi-label datasets can render traditional uncertainty-based selection methods less effective, resulting in suboptimal batch choices.


\section{Proposed Method}

\subsection{Preliminaries}

Let \(\mathcal{X} = \mathbb{R}^d\) denote the input space and \(\mathcal{Y} = \{l_1, l_2, \ldots, l_q\}\)
denote the label space with q labels.
A multi-label instance is denoted as \((\mathbf{x}_i, Y_i)\), with \(\mathbf{x}_i \in \mathcal{X}\) as the feature vector and \(Y_i \subseteq \mathcal{Y}\) indicating the relevant labels set. 
A binary vector \(\mathbf{y}_i = [y_{i1}, y_{i2}, \ldots, y_{iq}]\) of dimension \(q\), with elements in \(\{0, 1\}\), represents the set of labels \(Y_i\). Here, \(y_{ij} = 1\) implies that \(l_j \in Y_i\), and \(y_{ij} = 0\) indicates that \(l_j \notin Y_i\).
The objective of multi-label classification is to learn a prediction function \(f : \mathcal{X} \rightarrow 2^\mathcal{Y}\) based on a dataset \(D = \{(\mathbf{x}_i, Y_i) | 1 \leq i \leq n\}\). For any new example \(\mathbf{x}_u \in \mathcal{X}\), the function predicts a subset of labels \(f(\mathbf{x}_u) \subseteq \mathcal{Y}\) as relevant.

In multi-label data, the metric \(IRLbl\) \cite{MLSMOTE} is commonly used to evaluate the level of imbalance for a particular label. Let \(C_j^b\) be the count of samples for which the value of the \(j\)-th label is \(b \in \{0,1\}\).
Then the metric \(IRLbl\) is defined as:
\begin{equation}
IRLbl_j=C_{max}^1/C_j^1
\end{equation}
where \(C_{max}^1\) denotes the greatest number of instances for any label that is assigned a value of 1. 
$MeanIR = \frac{1}{q}\sum_{j=1}^{q}IRLbl_{j}$ measures the average imbalance across all labels. 
Let \(L_{m}\) represent the set of minority labels, defined by those labels whose \(IRLbl_{j}\) exceeds the \(MeanIR\). A label \(l_{j}\) is classified as a minority if \(IRLbl_{j} > MeanIR\) and as a majority otherwise. 

Let $\mathbf{B} \in \mathbb{R}^{n \times q}$ denote a local imbalance matrix, where $B_{ij}$ signifies the local imbalance for instance $\mathbf{x}_{i}$ regarding label $l_{j}$, defined by the proportion of neighboring instances that belong to a different class.
\begin{equation}
 B_{ij}=\left\{\begin{matrix}
\frac{1}{k}\sum_{(\mathbf{x}_{m},\mathbf{y}_{m})\in {K}_{\mathbf{x}_{i}}}  \llbracket y_{ij}\neq y_{mj}  \rrbracket,  & \quad y_{ij}=1 \\ 
0,  & otherwise 
\end{matrix}\right.
\label{localimb}
\end{equation}
where $\llbracket \pi  \rrbracket$ represents the indicator function, which yields 1 if $\llbracket \pi  \rrbracket$ is true and 0 otherwise, and $\mathcal{K}_{\mathbf{x}_{i}}$ represents the $k$-nearest neighbors ($k$NNs) of $\mathbf{x}_{i}$ based on the Euclidean distance.


\subsection{Rank-Based Batch Selection}

The rank-based batch selection \cite{Onlinebatchselection} sorts samples by loss, assigns sample selection probabilities based on rank, and selects mini-batches accordingly. Given a dataset of $n$ samples, the probability of selecting the $i$-th sample is:
\begin{equation}
p(i) = \frac{\exp\left(\log(s_e)/n\right)^{r(\ell_i)}}{\sum_{i=1}^{n} \exp\left(\log(s_e)/n\right)^{r(\ell_i)}}
\label{rankbatch}
\end{equation}
where \(r(\ell_i)\) denotes the rank of the elements in ascending order within \(\bm{\ell}\). Here \(\bm{\ell}\in \mathbb{R}^n\) is defined as the vector with its $i$-th element representing the model loss of the $i$-th sample. 
Furthermore, \(s_e\) represents the selection pressure, influencing the disparity in probabilities between the most and least significant samples.

In multi-label classification neural networks, the loss function usually involves several components. For example, the KL divergence is used by many models \cite{MPVAE,DELA,HOTVAE}, to quantify the difference between the distribution of input and latent space. However, this loss function component is not directly related to the classification target. We instead utilize multi-label binary cross-entropy as the loss for rank-based batch selection. The rationale behind this choice and its effectiveness are thoroughly discussed and validated in Section \ref{bceloss}.

\subsection{Class Imbalance Aware Weighting \label{insweight}}
To increase the likelihood of minority label instances being included in batches. we introduce a weight for each instance and incorporate it into  \(\bm{\ell}\). 
Let \(\mathbf{S} \in \mathbb{R}^{n\times q}\) represent the matrix that captures local imbalance, defined by:
\begin{equation}
S_{ij}=\left \{
\begin{aligned}
& \frac{B_{ij} \llbracket B_{ij} < 1 \rrbracket}{\sum_{i'=1}^{n} B_{i'j} \llbracket B_{i'j} < 1 \rrbracket} \quad  \text{if} \quad  \llbracket B_{i'j} < 1\rrbracket \\
& -1, \quad  \quad  \quad  \quad  \quad \quad \quad \quad\text{otherwise}
\end{aligned}
\right.
\label{eq:Sij}
\end{equation}
where normalization is applied to the local class imbalance of label $j$ for all instances to emphasize labels with fewer samples. 
Meanwhile, $\llbracket B_{i'j} < 1\rrbracket$ is used to exclude the influence of outliers. 
Further, let \(\bm{\epsilon} \in \mathbb{R}^n\) be the imbalance vector, with \(\epsilon_i = \sum_{l_j \in L_m} S_{ij} \llbracket S_{ij} \neq -1 \rrbracket\) for each instance, accumulating local imbalance from non-noisy minority labels.  
Given a vector \(\bm{\alpha}=\left [ 1,1,\cdots,1\right] \in \mathbb{R}^n\), the instance weight vector \(\mathbf{w} \in \mathbb{R}^n\) is defined as: \(\mathbf{w} = \bm{\alpha} + \bm{\epsilon}\).
Then, upon combining with \(\mathbf{w}\), the loss vector \(\bm{\ell}\) can be rewritten as:
\begin{equation}
\bm{\ell} = \mathbf{w} \odot \bm{\ell}.
\end{equation}
where $\odot$ represents the Adam product. 
It's important to highlight that \(\bm{\ell}\) is solely utilized for determining sample probability in the batch selection and does not contribute to the model's gradient update process. 
All data has a chance to be selected during the training process to avoid introducing system bias


\subsection{Incorporate Quantization Index}

In Eq.\ref{rankbatch}, the ranking-based method for sample selection probability tends to magnify small differences, resulting in significant (disproportionate) shifts in rankings and selection probabilities. 

A feasible way to circumvent this is to exploit the quantization index \cite{quantization} to smooth the rank value and limit it within the range of $n$. The quantization function $Q(\ell_i)$ \footnote{$\left\lceil \pi \right\rceil$ is upward rounding function} is given by:
\begin{equation} 
 Q(\ell_i)= \left\lceil \frac{\ell_i}{\Delta}\right\rceil 
\label{quantization}
\end{equation}
where \(\Delta\) is defined as the adaptive quantization step size, equivalent to \(\ell_{max}/n\). Here, \(\ell_{max}\) represents the maximum element of $\bm{\ell}$.
Combining $Q(\ell_i)$, the probability of sample selection in multi-label adaptive batch can be rewritten as:
\begin{equation}
p(i) = \frac{\exp\left(\log(s_e)/n\right)^{Q(\ell_i)}}{\sum_{j=1}^{n} \exp\left(\log(s_e)/n\right)^{Q(\ell_j)}}
\label{samplingprobability}
\end{equation}
\begin{figure}
    \centering \includegraphics[width=\textwidth]{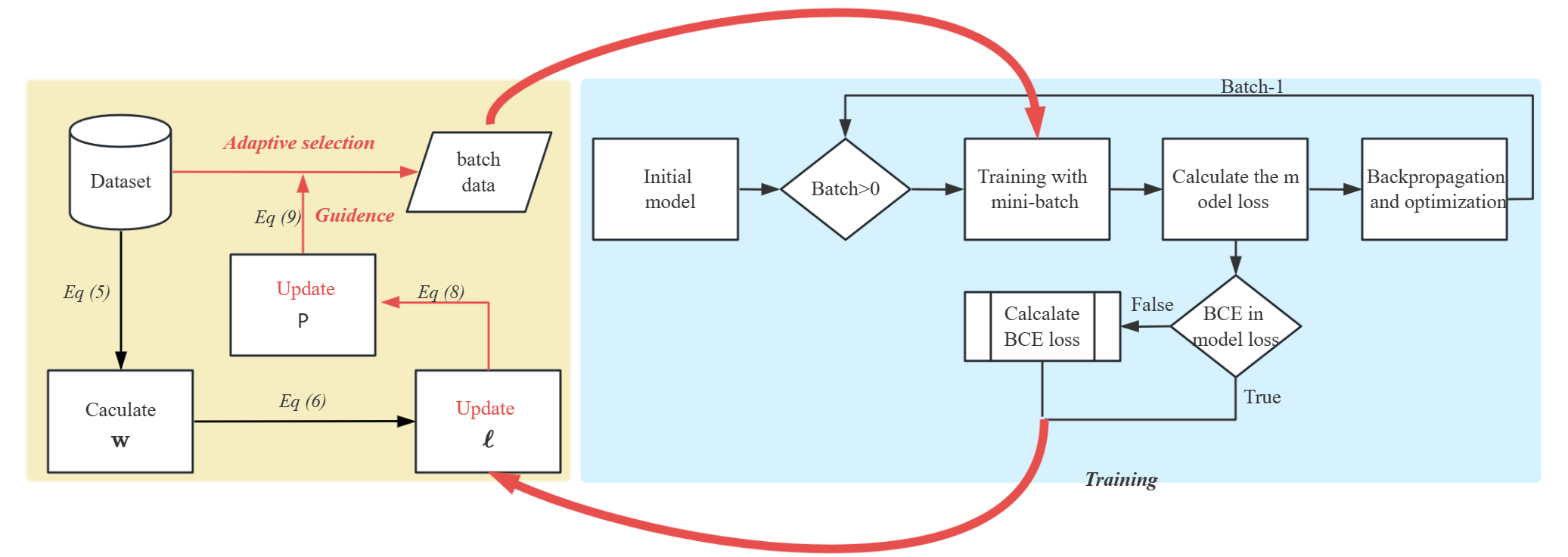}
    \caption{The work flow of multi-label deep learning model training with adaptive batch selection}
    \label{fig:flowchart}
\end{figure}

Figure \ref{fig:flowchart} shows the workflow of the model training with adaptive batch selection, the red line indicates adaptability on a batch basis.

Before the training, the set $P$ is initialized to track each sample's $p(i)$. 
Meanwhile, the model requires a continuous warm-up phase of $\gamma$ epochs to alleviate the instability caused by initial random initialization.
In adaptive batch selection, each sample's selection probability is linked to its quantization index \(q\), calculated by \(\bm{\ell}\) using an adaptive step size \(\Delta\). Here, \(\ell_{max}\) denotes the highest loss among all samples after the previous batch of training ends. After each batch, we adjust the quantization index based on \(\bm{\ell}\) and then update \(P\) by recalculating each sample's $p(i)$.

\subsection{Variant of Adaptive Batch Selection Exploiting Label Correlations}
It is well-known that label correlations are significant in multi-label learning. Label correlations can provide additional information, especially when some labels have insufficient training samples. 
By considering label correlation, we propose a chain-based adaptive batch selection method. 
First, we select the seed sample based on the selection probability set $P$. 
Given that $l_j$ is the seed sample's associated label with the maximum $IRLbl$, we employ the adjacency matrix $\mathbf{A}$ to identify the $\left\lceil Card \right\rceil$ ($Card$ represents the label cardinality) labels most closely related to $l_j$ \footnote{\(\mathbf{A} \in \mathbb{R}^{q\times q}\) is the symmetric conditional probability matrix, which measures the co-occurrence relationship between labels, and the definition can be found in the appendix of supplementary materials.}, constituting the set $L_c$. We then define $D_c$ as the collection of instances linked to the labels within $L_c$.
From the second sample onward, selection relies on a new probability set $P_c$, defined as $P_{cor}=\left \{p(i) \mid (\mathbf{x}_i,\mathbf{y_{i}}) \in D_c \right \}$.
The selected sample then becomes the new seed sample for the next round of selection. This process is repeated until it reaches the batch size.\\


\subsection{Convergence Guarantee}
Adaptive batch selection methods typically meet Adam's convergence standards, assuming their sampling distributions are \textbf{strictly positive} and \textbf{gradient estimates unbiased}. We theoretically demonstrate that our approach meets these key convergence conditions and the \textbf{Proof}  can be found in the appendix of supplementary materials. 



\section{Experiments and Analysis}
\subsection{Experiment Setup}
\subsubsection{Datasets}
We analyze thirteen multi-label datasets spanning text, image, and bioinformatics domains, sourced from the MULAN repository \cite{MULAN}.  
\begin{table}[!h]
\centering
\caption{The multi-label datasets used in the experiments.} 
\label{ta:mld}
\resizebox{\textwidth}{!}{
\begin{tabular}{cccccccccccccccc}
\toprule
name & $n$ & $d$ & $q$ & Card & Dens & MeanIR & domain & name & $n$ & $d$ & $q$ & Card & Dens & MeanIR & domain\\  
\midrule
Corel5k\textsuperscript{a} & 5000 & 499 & 374 & 3.52 & 0.01 & 189.57 & images& tmc2007& 28596 & 490 & 22 & 2.15 & 0.10 & 15.16 & text \\
rcv1subset3 & 6000 & 944 & 101 & 2.61 & 0.03 & 68.33 & text& bibtex & 7395 & 1836 & 159 & 2.40 & 0.02 & 12.50 & text \\
rcv1subset1 & 6000 & 944 & 101 & 2.88 & 0.03 & 54.49 & text & enron & 1702 & 1001 & 53 & 3.38 & 0.06 & 9.93 & text \\
rcv1subset2 & 6000 & 944 & 101 & 2.63 & 0.03 & 45.51 & text& yeast & 2417 & 103 & 14 & 4.24 & 0.30 & 7.20 & biology \\
yahoo-Arts & 7484 & 2314 & 25 & 1.67 & 0.07 & 26.00 & text & LLOG-F & 1460 & 1004 & 75 & 15.93 & 0.21 & 5.39 & text \\
cal500 & 502 & 68 & 174 & 26.04 & 0.15 & 20.58 & music & scene & 2407 & 294 & 6 & 1.07 & 0.18 & 1.25 & images \\
yahoo-Business & 11214 & 2192 & 28 & 1.47 & 0.06 & 16.90 & text \\
\bottomrule
\end{tabular}
}
\end{table}
Characteristics and imbalance levels of these datasets are detailed in Table \ref{ta:mld}, including Card, the mean labels per instance associated, and Dens, the ratio of Card to the overall label count.

\subsubsection{Evaluation Metrics}
To assess the efficacy of the batch method in multi-label classification, six commonly utilized evaluation metrics are adopted, comprising Macro-F, Micro-F, Macro-AUC, Ranking Loss, Hamming Loss, and One Error. Please refer to \cite{mlcreview1,mlcreview2} for detailed definitions of these metrics.

\subsubsection{Base Classifier and Implementation Details}
We used five multi-label deep models, namely C2AE \cite{C2AE}, MPVAE \cite{MPVAE}, PACA \cite{PACA}, CLIF \cite{CLIF}, and DELA \cite{DELA}. 
We configure each model according to the parameter settings provided in the original paper and its source code, including layer dimensions, activation functions, and other specifics. For any hyperparameters not detailed, we standardize configurations across batch selection methods to maintain consistency.
For optimization, Adam, with a batch size of 128, weight decay of 1e-4, and momentums of 0.999 and 0.9, is employed. 
As for the adaptive batch selection parameters, we used the best selection pressure $s_e$, obtained from $s_e = \{2,8,16,64\}$, and set the warm-up threshold $\gamma$ to 3. Technically, a small $\gamma$ is enough to warm up, but to reduce the performance variance caused by randomly initialized, we use the first three epochs as the warm-up period and share the model parameters for all strategies during the warm-up period \cite{Ada-boundary}. 
We employ five-fold cross-validation to evaluate the above approaches on the 13 data sets. In each fold, we record the results on the test set at the best epoch of the validation set. 
Our code can be found in Anonymous GitHub \href{https://anonymous.4open.science/r/MLAdaBatch}{1}.

\subsection{Experimental Results}

Table \ref{tab:result} shows the comparative average performance metrics of two different batch selection techniques, adaptive batch selection, and random batch selection, in different models.
"Random" denotes the use of random batch selection with "shuffle=True", which refers to the batch selection used by these models. 
Adaptive batch selection consistently outperformed random selection, achieving the highest rankings across various evaluation metrics in most datasets. This advantage was particularly pronounced in datasets with high imbalance, such as Corel5k, rcv1subset1, rcv1subset2, rcv1subset3, as well as in those with a large number of labels, such as Corel5k, bibtex, and cal500. 
Adaptive batch selection significantly enhances the performance of embedding-based models like C2AE and models employing sophisticated neural networks, such as the CLIF. These results highlight our method's broad effectiveness in deep learning models. 
\begin{table}[!ht]
\centering
\resizebox{0.95\textwidth}{!}{
\begin{tabular}{cccccccccccccc}
\toprule
\multirow{16}{*}{\textbf{C2AE}} &  & \multicolumn{2}{c}{Macro-F$\uparrow$} & \multicolumn{2}{c}{Micro-F$\uparrow$} & \multicolumn{2}{c}{Macro-AUC$\uparrow$} & \multicolumn{2}{c}{Ranking Loss$\downarrow$} & \multicolumn{2}{c}{Hamming Loss$\downarrow$} & \multicolumn{2}{c}{One Error$\downarrow$} \\ \midrule
 & Dataset & Random & Adaptive & Random & Adaptive & Random & Adaptive & Random & Adaptive & Random & Adaptive & Random & Adaptive \\
 & Corel5k & 0.0702 & \textbf{0.0726} & 0.1943 & \textbf{0.1985} & 0.6165 & \textbf{0.6361} & 0.2922 & \textbf{0.2832} & 0.0533 & \textbf{0.0393} & 0.8202 & \textbf{0.8002} \\
 & rcv1subset3 & 0.2659 & \textbf{0.2768} & 0.4387 & \textbf{0.4481} & 0.8136 & \textbf{0.8298} & 0.1287 & \textbf{0.1187} & 0.0362 & \textbf{0.0344} & 0.5423 & \textbf{0.5321} \\
 & rcv1subset1 & 0.2611 & 0.2723 & 0.4565 & \textbf{0.4664} & 0.8396 & \textbf{0.8589} & 0.0930 & \textbf{0.0917} & 0.0399 & \textbf{0.0380} & 0.5081 & \textbf{0.4918} \\
 & rcv1subset2 & 0.2505 & \textbf{0.2609} & 0.4463 & \textbf{0.4544} & 0.8157 & \textbf{0.8277} & 0.1114 & \textbf{0.1067} & 0.0369 & \textbf{0.0346} & 0.5505 & \textbf{0.5355} \\
 & yahoo-Arts1 & 0.2007 & \textbf{0.2146} & 0.3973 & 0.4064 & 0.7108 & \textbf{0.7287} & 0.1188 & \textbf{0.1129} & \textbf{0.0528} & \textbf{0.0528} & 0.5192 & \textbf{0.5147} \\
 & cal500 & 0.2473 & \textbf{0.2634} & 0.4641 & \textbf{0.4880} & 0.5781 & \textbf{0.5972} & 0.2610 & \textbf{0.2413} & 0.2616 & \textbf{0.2417} & 0.1615 & \textbf{0.1391} \\
 & yahoo-Business1 & 0.2224 & \textbf{0.2243} & 0.4152 & \textbf{0.4176} & \textbf{0.7478} & 0.7455 & \textbf{0.0373} & 0.0379 & 0.0452 & \textbf{0.0446} & 0.8239 & \textbf{0.8199} \\
 & tmc2007 & 0.4332 & \textbf{0.4405} & 0.5643 & \textbf{0.5895} & 0.8697 & \textbf{0.8711} & 0.0785 & \textbf{0.0764} & 0.0700 & \textbf{0.0685} & 0.2710 & \textbf{0.2691} \\
 & bibtex & 0.2683 & \textbf{0.2778} & 0.3736 & \textbf{0.3788} & 0.8594 & \textbf{0.8674} & 0.1318 & \textbf{0.1216} & 0.0194 & \textbf{0.0147} & 0.4878 & \textbf{0.4443} \\
 & enron & 0.2587 & 0.2692 & 0.5510 & 0.5622 & 0.6690 & \textbf{0.6698} & 0.1610 & \textbf{0.1496} & 0.0901 & \textbf{0.0840} & 0.3771 & \textbf{0.3354} \\
 & yeast & 0.4040 & \textbf{0.4275} & 0.6536 & \textbf{0.6567} & 0.7045 & \textbf{0.7100} & \textbf{0.1719} & 0.1723 & 0.2011 & \textbf{0.1981} & 0.2527 & \textbf{0.2362} \\
 & LLOG-F & 0.3021 & \textbf{0.3066} & 0.5556 & \textbf{0.5694} & 0.6375 & \textbf{0.6419} & 0.2011 & \textbf{0.1961} & 0.2075 & \textbf{0.1976} & 0.2340 & 0.2066 \\
 & scene & 0.7249 & \textbf{0.7368} & 0.7163 & \textbf{0.7277} & 0.9303 & \textbf{0.9361} & 0.0920 & \textbf{0.0892} & 0.0968 & \textbf{0.0942} & 0.2605 & \textbf{0.2508} \\ \midrule
\multirow{16}{*}{\textbf{MPVAE}} &  & \multicolumn{2}{c}{Macro-F$\uparrow$} & \multicolumn{2}{c}{Micro-F$\uparrow$} & \multicolumn{2}{c}{Macro-AUC$\uparrow$} & \multicolumn{2}{c}{Ranking Loss$\downarrow$} & \multicolumn{2}{c}{Hamming Loss$\downarrow$} & \multicolumn{2}{c}{One Error$\downarrow$} \\ \midrule
 & Corel5k & 0.1131 & \textbf{0.1155} & 0.2196 & \textbf{0.2311} & 0.7025 & \textbf{0.7054} & \textbf{0.2316} & 0.2322 & 0.0254 & \textbf{0.0245} & \textbf{0.6903} & 0.7009 \\
 & rcv1subset3 & 0.3190 & \textbf{0.3262} & 0.4733 & \textbf{0.4751} & 0.8971 & \textbf{0.9051} & 0.0702 & \textbf{0.0697} & 0.0301 & \textbf{0.0296} & 0.4207 & \textbf{0.4182} \\
 & rcv1subset1 & 0.3514 & \textbf{0.3605} & 0.4664 & \textbf{0.4780} & 0.8997 & \textbf{0.9071} & 0.0693 & \textbf{0.0681} & 0.0358 & \textbf{0.0349} & 0.4275 & \textbf{0.4148} \\
 & rcv1subset2 & 0.3532 & \textbf{0.3595} & 0.4820 & \textbf{0.4854} & 0.9032 & \textbf{0.9082} & 0.0659 & \textbf{0.0628} & 0.0302 & 0.0300 & 0.4173 & \textbf{0.4112} \\
 & yahoo-Arts1 & 0.3364 & \textbf{0.3528} & 0.4684 & \textbf{0.4780} & 0.7456 & \textbf{0.7492} & 0.1384 & \textbf{0.1341} & 0.0572 & \textbf{0.0565} & 0.5095 & \textbf{0.5039} \\
 & cal500 & 0.2026 & \textbf{0.2084} & 0.3664 & \textbf{0.3783} & 0.5365 & \textbf{0.5379} & \textbf{0.3183} & 0.3205 & 0.2394 & \textbf{0.2381} & 0.3662 & \textbf{0.3648} \\
 & yahoo-Business1 & 0.3409 & \textbf{0.3580} & 0.4670 & \textbf{0.4774} & 0.7865 & \textbf{0.7923} & 0.0452 & \textbf{0.0421} & 0.0179 & \textbf{0.0161} & 0.8235 & \textbf{0.8202} \\
 & tmc2007 & 0.5154 & \textbf{0.5242} & 0.6275 & \textbf{0.6326} & 0.8918 & \textbf{0.8947} & 0.0612 & \textbf{0.0606} & 0.0700 & 0.0696 & 0.2583 & \textbf{0.2528} \\
 & bibtex & \textbf{0.3395} & 0.3385 & 0.4522 & \textbf{0.4543} & 0.8724 & \textbf{0.8740} & 0.1045 & \textbf{0.1026} & 0.0145 & \textbf{0.0144} & 0.4101 & \textbf{0.4041} \\
 & enron & \textbf{0.3007} & 0.2964 & 0.5531 & \textbf{0.5548} & \textbf{0.7251} & 0.7241 & 0.1418 & \textbf{0.1390} & 0.0838 & \textbf{0.0832} & 0.3041 & \textbf{0.2967} \\
 & yeast & 0.4434 & \textbf{0.4486} & 0.6235 & \textbf{0.6261} & 0.6949 & \textbf{0.7013} & 0.1933 & \textbf{0.1907} & 0.2178 & 0.2182 & \textbf{0.2708} & 0.2710 \\
 & LLOG-F & 0.4179 & \textbf{0.4297} & 0.5531 & \textbf{0.5635} & 0.7667 & \textbf{0.7692} & 0.2387 & \textbf{0.2360} & 0.1915 & \textbf{0.1908} & 0.3070 & \textbf{0.2796} \\
 & scene & 0.7702 & \textbf{0.7833} & 0.7608 & \textbf{0.7762} & 0.9462 & \textbf{0.9484} & 0.0693 & \textbf{0.0655} & 0.0835 & \textbf{0.0782} & 0.2106 & \textbf{0.1965} \\ \midrule
\multirow{16}{*}{\textbf{PACA}} &  & \multicolumn{2}{c}{Macro-F$\uparrow$} & \multicolumn{2}{c}{Micro-F$\uparrow$} & \multicolumn{2}{c}{Macro-AUC$\uparrow$} & \multicolumn{2}{c}{Ranking Loss$\downarrow$} & \multicolumn{2}{c}{Hamming Loss$\downarrow$} & \multicolumn{2}{c}{One Error$\downarrow$} \\ \midrule
 & Corel5k & \textbf{0.1447} & 0.1431 & \textbf{0.2372} & 0.2359 & 0.7468 & \textbf{0.7543} & 0.1618 & \textbf{0.1587} & 0.0282 & \textbf{0.0277} & \textbf{0.6982} & 0.7047 \\
 & rcv1subset3 & 0.3500 & \textbf{0.3576} & 0.4870 & \textbf{0.4899} & 0.9091 & \textbf{0.9165} & 0.0526 & \textbf{0.0507} & 0.0314 & \textbf{0.0306} & 0.4125 & \textbf{0.4077} \\
 & rcv1subset1 & 0.3540 & \textbf{0.3586} & 0.4587 & \textbf{0.4618} & 0.9066 & 0.9099 & 0.0528 & \textbf{0.0515} & 0.0367 & \textbf{0.0364} & 0.4301 & \textbf{0.4243} \\
 & rcv1subset2 & 0.3512 & \textbf{0.3547} & \textbf{0.4893} & 0.4856 & 0.9067 & 0.9158 & 0.0520 & \textbf{0.0512} & 0.0310 & \textbf{0.0307} & 0.4138 & \textbf{0.4098} \\
 & yahoo-Arts1 & 0.2273 & \textbf{0.2331} & 0.3981 & \textbf{0.4032} & 0.7523 & \textbf{0.7612} & 0.1365 & \textbf{0.1314} & 0.0524 & \textbf{0.0515} & 0.5473 & \textbf{0.5408} \\
 & cal500 & 0.1215 & \textbf{0.1286} & 0.3705 & \textbf{0.3721} & 0.5821 & \textbf{0.5881} & 0.2392 & \textbf{0.2345} & 0.1957 & \textbf{0.1930} & 0.1179 & \textbf{0.1157} \\
 & yahoo-Business1 & \textbf{0.3219} & 0.3207 & 0.4940 & \textbf{0.5041} & 0.8236 & \textbf{0.8279} & \textbf{0.0349} & 0.0368 & 0.0155 & \textbf{0.0134} & 0.8119 & \textbf{0.8091} \\
 & tmc2007 & 0.4890 & \textbf{0.4902} & 0.6040 & \textbf{0.6132} & 0.8837 & \textbf{0.8849} & 0.0645 & \textbf{0.0642} & 0.0749 & \textbf{0.0743} & 0.2778 & \textbf{0.2726} \\
 & bibtex & 0.2995 & \textbf{0.3008} & 0.4101 & \textbf{0.4128} & 0.8476 & \textbf{0.8528} & \textbf{0.0923} & 0.0926 & 0.0157 & \textbf{0.0128} & 0.4524 & \textbf{0.4497} \\
 & enron & 0.3051 & \textbf{0.3071} & 0.5762 & \textbf{0.5781} & 0.7351 & \textbf{0.7391} & 0.1135 & \textbf{0.1095} & \textbf{0.0845} & \textbf{0.0845} & 0.2801 & \textbf{0.2761} \\
 & yeast & 0.4180 & \textbf{0.4232} & 0.6435 & \textbf{0.6476} & 0.7096 & \textbf{0.7097} & 0.1737 & \textbf{0.1721} & 0.2051 & \textbf{0.2010} & 0.2391 & \textbf{0.2382} \\
 & LLOG-F & 0.3837 & \textbf{0.3996} & 0.5612 & \textbf{0.5643} & 0.7104 & \textbf{0.7131} & 0.2022 & \textbf{0.1968} & 0.1791 & \textbf{0.1783} & 0.2365 & \textbf{0.2318} \\
 & scene & 0.7712 & \textbf{0.7727} & 0.7640 & \textbf{0.7697} & 0.9442 & \textbf{0.9464} & 0.0663 & \textbf{0.0667} & 0.0843 & \textbf{0.0827} & 0.2015 & \textbf{0.2011} \\\midrule
\multirow{16}{*}{\textbf{CLIF}} &  & \multicolumn{2}{c}{Macro-F$\uparrow$} & \multicolumn{2}{c}{Micro-F$\uparrow$} & \multicolumn{2}{c}{Macro-AUC$\uparrow$} & \multicolumn{2}{c}{Ranking Loss$\downarrow$} & \multicolumn{2}{c}{Hamming Loss$\downarrow$} & \multicolumn{2}{c}{One Error$\downarrow$} \\ \midrule
 & Corel5k & 0.1246 & \textbf{0.1269} & 0.2375 & \textbf{0.2379} & 0.7534 & \textbf{0.7683} & \textbf{0.1677} & 0.1762 & 0.0241 & \textbf{0.0240} & \textbf{0.6628} & 0.6660 \\
 & rcv1subset3 & 0.3057 & \textbf{0.3079} & 0.4729 & \textbf{0.4738} & 0.9268 & \textbf{0.9279} & 0.0606 & \textbf{0.0602} & 0.0341 & \textbf{0.0294} & 0.4129 & \textbf{0.4113} \\
 & rcv1subset1 & 0.3498 & \textbf{0.3506} & 0.4711 & \textbf{0.4746} & 0.9221 & \textbf{0.9307} & 0.0617 & \textbf{0.0576} & 0.0355 & \textbf{0.0338} & 0.4192 & \textbf{0.4072} \\
 & rcv1subset2 & 0.3265 & \textbf{0.3284} & 0.4756 & \textbf{0.4824} & 0.9279 & \textbf{0.9329} & 0.0611 & \textbf{0.0610} & 0.0302 & \textbf{0.0299} & 0.4272 & \textbf{0.4166} \\
 & yahoo-Arts1 & 0.3384 & \textbf{0.3412} & 0.4772 & \textbf{0.4797} & \textbf{0.7580} & 0.7535 & 0.1293 & \textbf{0.1207} & 0.0556 & \textbf{0.0561} & 0.4936 & \textbf{0.4928} \\
 & cal500 & 0.1007 & \textbf{0.1066} & 0.3547 & \textbf{0.3620} & 0.5801 & \textbf{0.5982} & 0.2300 & \textbf{0.2259} & 0.1910 & \textbf{0.1895} & 0.1215 & \textbf{0.1196} \\
 & yahoo-Business1 & \textbf{0.3765} & 0.3761 & \textbf{0.5097} & 0.5082 & 0.7801 & \textbf{0.7940} & 0.0430 & \textbf{0.0374} & \textbf{0.0163} & 0.0165 & 0.8099 & \textbf{0.8087} \\
 & tmc2007 & \textbf{0.5337} & 0.5306 & \textbf{0.6451} & 0.6355 & 0.9048 & \textbf{0.9059} & 0.0552 & \textbf{0.0550} & 0.0700 & \textbf{0.0697} & 0.2540 & \textbf{0.2521} \\
 & bibtex & 0.3423 & \textbf{0.3467} & 0.4687 & \textbf{0.4696} & 0.8983 & \textbf{0.9011} & 0.0884 & \textbf{0.0841} & 0.0133 & \textbf{0.0132} & \textbf{0.3767} & 0.3846 \\
 & enron & 0.2903 & \textbf{0.2909} & 0.5728 & \textbf{0.5770} & 0.7700 & \textbf{0.7763} & 0.1232 & \textbf{0.1198} & 0.0758 & \textbf{0.0745} & 0.2504 & \textbf{0.2492} \\
 & yeast & 0.4141 & \textbf{0.4157} & 0.6482 & \textbf{0.6561} & 0.7107 & \textbf{0.7191} & 0.1679 & \textbf{0.1607} & 0.1995 & \textbf{0.1930} & 0.2429 & \textbf{0.2221} \\
 & LLOG-F & 0.3962 & \textbf{0.3958} & \textbf{0.5671} & 0.5661 & 0.7659 & \textbf{0.7703} & 0.2169 & \textbf{0.2138} & 0.2000 & \textbf{0.1691} & 0.2340 & \textbf{0.2326} \\
 & scene & 0.7606 & \textbf{0.7709} & 0.7473 & \textbf{0.7589} & 0.9418 & \textbf{0.9454} & 0.0725 & \textbf{0.0676} & 0.0891 & \textbf{0.0850} & 0.2284 & \textbf{0.2093} \\ \midrule
\multirow{16}{*}{\textbf{DELA}} &  & \multicolumn{2}{c}{Macro-F$\uparrow$} & \multicolumn{2}{c}{Micro-F$\uparrow$} & \multicolumn{2}{c}{Macro-AUC$\uparrow$} & \multicolumn{2}{c}{Ranking Loss$\downarrow$} & \multicolumn{2}{c}{Hamming Loss$\downarrow$} & \multicolumn{2}{c}{One Error$\downarrow$} \\ \midrule
 & Corel5k & 0.0755 & \textbf{0.0963} & 0.1812 & \textbf{0.2081} & 0.7556 & \textbf{0.7621} & 0.1601 & \textbf{0.1532} & 0.0216 & \textbf{0.0223} & 0.6526 & \textbf{0.6520} \\
 & rcv1subset3 & 0.2673 & \textbf{0.2995} & 0.4640 & \textbf{0.4704} & 0.9175 & \textbf{0.9183} & 0.0569 & \textbf{0.0581} & 0.0305 & \textbf{0.0300} & 0.4307 & \textbf{0.4080} \\
 & rcv1subset1 & 0.2842 & \textbf{0.3184} & 0.4424 & \textbf{0.4544} & 0.9179 & \textbf{0.9194} & 0.0576 & \textbf{0.0571} & 0.0362 & \textbf{0.0356} & 0.4355 & \textbf{0.4305} \\
 & rcv1subset2 & 0.2868 & \textbf{0.3125} & 0.4680 & \textbf{0.4748} & 0.9203 & \textbf{0.9220} & 0.0540 & \textbf{0.0538} & 0.0312 & \textbf{0.0303} & 0.4358 & \textbf{0.4118} \\
 & yahoo-Arts1 & \textbf{0.2735} & 0.2688 & 0.4617 & \textbf{0.4628} & 0.7387 & \textbf{0.7444} & 0.1227 & \textbf{0.1202} & 0.0547 & \textbf{0.0537} & \textbf{0.5131} & 0.5136 \\
 & cal500 & 0.0666 & \textbf{0.0833} & 0.3283 & \textbf{0.3473} & 0.5633 & \textbf{0.5927} & 0.2322 & \textbf{0.2273} & 0.1904 & \textbf{0.1894} & 0.1215 & \textbf{0.1156} \\
 & yahoo-Business1 & 0.3295 & \textbf{0.3196} & \textbf{0.4942} & 0.4913 & 0.7979 & \textbf{0.8062} & 0.0358 & \textbf{0.0343} & 0.0173 & \textbf{0.0167} & \textbf{0.8119} & 0.8153 \\
 & tmc2007 & 0.5417 & \textbf{0.5495} & 0.6546 & \textbf{0.6550} & 0.9121 & \textbf{0.9171} & 0.0524 & \textbf{0.0517} & 0.0673 & \textbf{0.0667} & 0.2544 & \textbf{0.2504} \\
 & bibtex & 0.2956 & \textbf{0.3151} & 0.4399 & \textbf{0.4517} & 0.9042 & \textbf{0.9053} & 0.0767 & \textbf{0.0746} & 0.0133 & \textbf{0.0137} & 0.4024 & \textbf{0.4012} \\
 & enron & \textbf{0.2743} & 0.2688 & 0.5800 & \textbf{0.5931} & 0.7727 & \textbf{0.7748} & 0.1154 & \textbf{0.1108} & 0.0735 & \textbf{0.0722} & 0.2421 & 0.2426 \\
 & yeast & 0.3813 & \textbf{0.3859} & 0.6482 & \textbf{0.6551} & 0.7006 & \textbf{0.7121} & 0.1677 & \textbf{0.1631} & 0.1975 & \textbf{0.1933} & 0.2328 & \textbf{0.2160} \\
 & LLOG-F & 0.3749 & \textbf{0.3798} & 0.5208 & \textbf{0.5859} & 0.7909 & \textbf{0.7930} & 0.1865 & \textbf{0.1849} & 0.1603 & \textbf{0.1601} & \textbf{0.1752} & 0.1806 \\
 & scene & 0.7496 & \textbf{0.7684} & 0.7396 & \textbf{0.7585} & 0.9405 & \textbf{0.9449} & 0.0722 & 0.0726 & 0.0898 & \textbf{0.0834} & 0.2309 & \textbf{0.2183} \\ \bottomrule
\end{tabular}
}
\caption{Results of batch selection strategies using five MLC Deep classifier with Adam optimizer on different datasets}
\label{tab:result}
\end{table}

The results presented in Table \ref{tab:Wilcoxon}, derived from the Wilcoxon signed-ranks test \cite{Wilcoxonsigned-rank} at a 0.05 significance level, conclusively indicate that our adaptive batch selection method outperforms random batch selection with statistical significance.

\begin{table}[!h]
\resizebox{0.8\textwidth}{!}{
\begin{tabular}{cccccc}
\toprule
 & C2AE & MPVAE & PACA & CLIF & DELA \\ \midrule
Macro-F & \textbf{win} (0.0002) & \textbf{win} (0.0017) & \textbf{win} (0.0046) & \textbf{win} (0.0215)  & \textbf{win} (0.0171) \\
Micro-F & \textbf{win} (0.0002) & \textbf{win} (0.0002) & \textbf{win} (0.0081) & tie (0.0942) & \textbf{win} (0.0012) \\
Macro-AUC & \textbf{win} (0.0012) & \textbf{win} (0.0005) & \textbf{win} (0.0002) & \textbf{win} (0.0034) & \textbf{win} (0.0002) \\
Ranking Loss & \textbf{win} (0.0012) & \textbf{win} (0.0061) & \textbf{win} (0.0171) & \textbf{win} (0.0171) & \textbf{win} (0.0046) \\
Hamming Loss & \textbf{win} (0.0022) & \textbf{win} (0.0012) & \textbf{win} (0.0022) & \textbf{win} (0.0081) & \textbf{win} (0.0105) \\
One Error & \textbf{win} (0.0022) & \textbf{win} (0.0134) & \textbf{win} (0.0171) & \textbf{win} (0.0479) & tie (0.1465) \\ \bottomrule
\end{tabular}
}
\caption{Summary of the Wilcoxon Signed-Ranks Test for adaptive against random in terms of each evaluation metric at 0.05 significance level. p-values are shown in the brackets}
\label{tab:Wilcoxon}
\end{table}


\subsection{Convergence Analysis \label{convergencecurve}}

\begin{figure}[!h]
    \centering
    \subfigure[Epoch perspective]{\includegraphics[width=\textwidth]{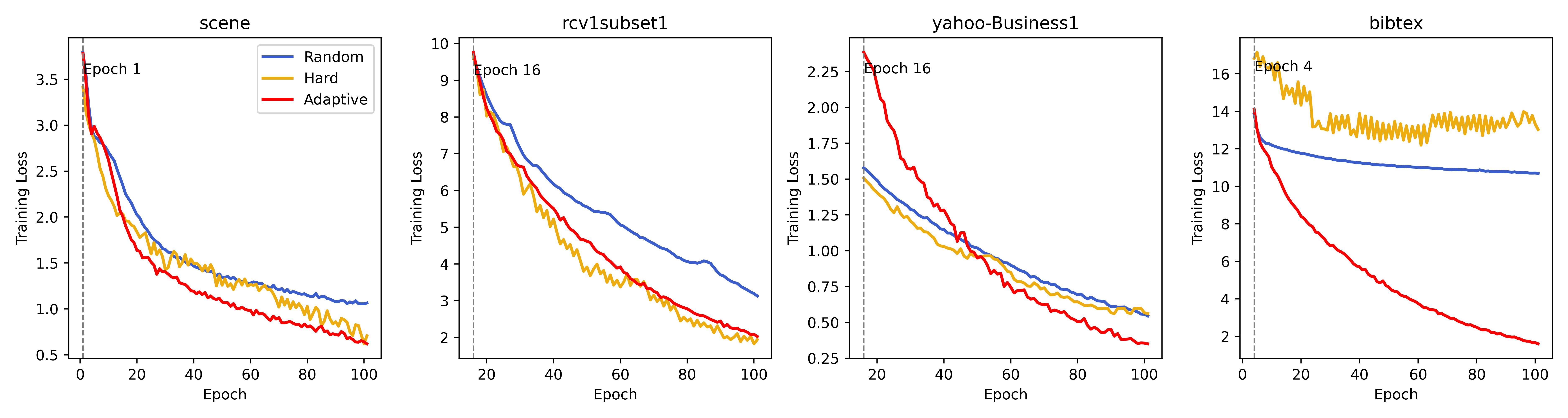}
    \label{fig:epochconvergence}
    }
    \subfigure[Batch perspective]{\includegraphics[width=\textwidth]{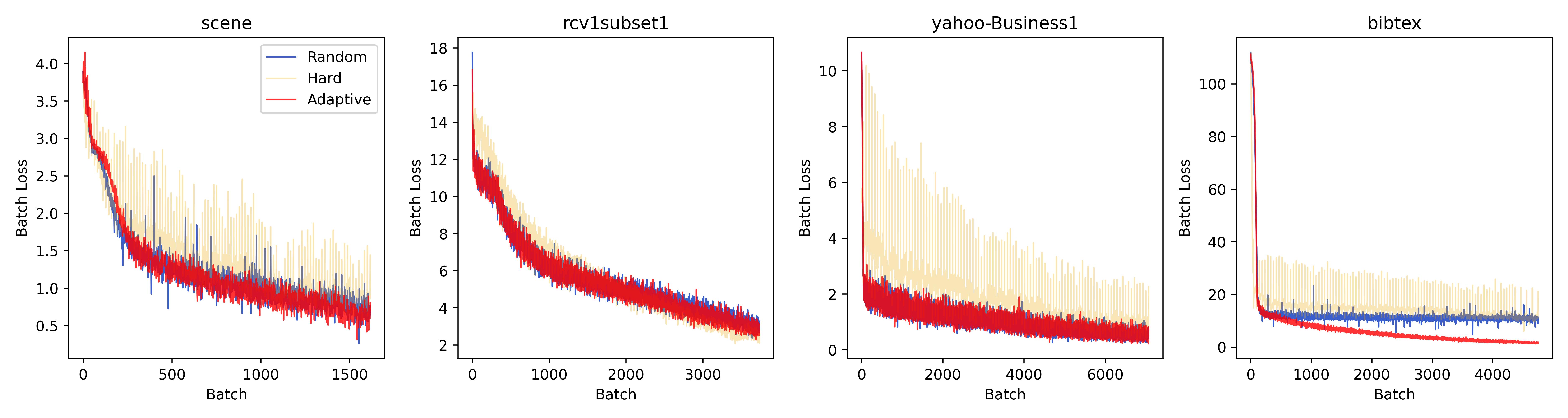}
    \label{fig:batchconvergence}
    } 
    \subfigure[Time perspective]{\includegraphics[width=\textwidth]{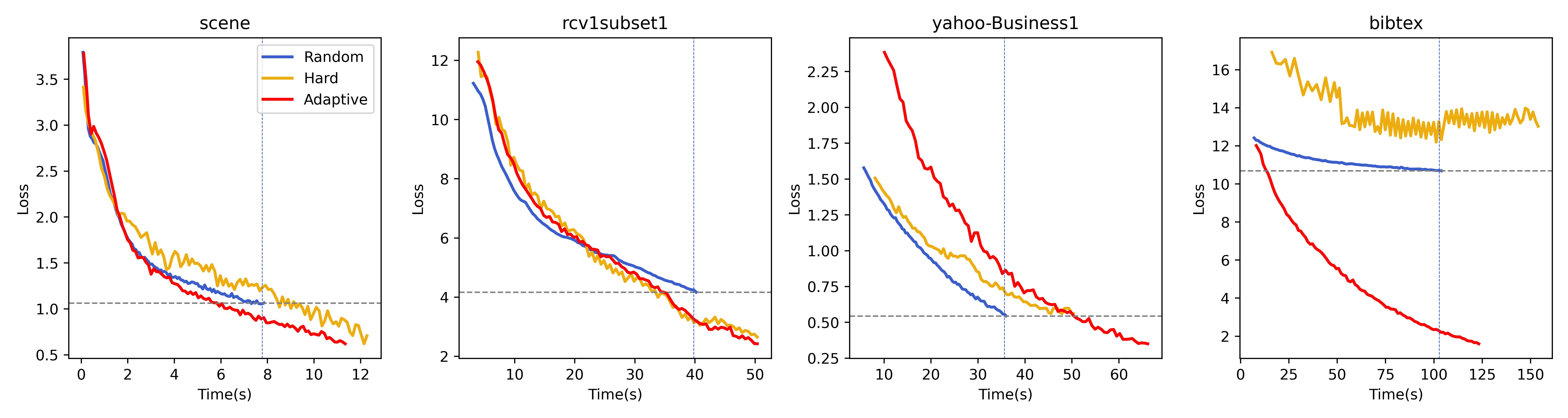}
    \label{fig:timeconvergence}
    } 
    \caption{Convergence curves of three batch selection strategies using DELA with Adam optimizer on different datasets.}
    \label{fig:convergence}
\end{figure}

In this section, we add a ranking batch selection based on BCE loss (named "Hard") as a comparison from the ablation perspective. Compared with the Adaptive method, Hard ignores imbalance weight $\mathbf{w}$ and does not employ the quantization index.
Figure \ref{fig:convergence} illustrates the convergence curves of three distinct batch selection strategies, viewed from the perspectives of epochs, batches, and time, along with the validation set performance curves evaluated using Macro-AUC. 

\textbf{Epoch Perspective}: As shown in Figure \ref{fig:epochconvergence}, the adaptive batch selection outperforms the other strategies in the scene and bibtex dataset, converging to lower losses with fewer epochs. Hard batch selection method shows improved convergence over random batch methods on the first three datasets but underperforms on the bibtex datatset.  
    The latter may be due to the small difference in training sample loss, failing the loss ranking strategy between samples. The convergence curve of the adaptive batch selection proves the effectiveness of the quantization index.
    Specifically, we observe that training with random batch, the loss on the bibtex dataset plateaued after only a few epochs. In contrast, adopting adaptive batching allowed the model to break through this "bottleneck", leading to a continued decline in the loss convergence curve.
    
    \textbf{Batch Perspective}: As shown in Figure \ref{fig:batchconvergence}, the adaptive batch shows less volatility and a steadier loss reduction during training. In contrast, the hard batch method is more volatile. The random batch method displays moderate volatility. The stability offered by adaptive batch selection is advantageous since large fluctuations, particularly upward ones in the last batch of an epoch, could disrupt the entire training epoch and adversely affect validation performance.
    
     \textbf{Time Perspective}: As shown in Figure \ref{fig:timeconvergence}, to evaluate performance gains in training time, we measure the time required to achieve the same training loss levels (based on the minimum training loss achieved by a random batch). For example, on the scene dataset, random batch selection requires approximately 7.8 seconds to reach the minimum loss, while the adaptive batch selection only needs 5.7 seconds. On the scene, rcv1subset1, and bibtex datasets, the adaptive batch reduces training time by 14.1$\%$, 10.8$\%$, and 84.6$\%$, respectively. Although the adaptive batch increases training time by 37.2$\%$ on the yahoo-Business dataset, the overall convergence curve and evaluation metric results suggest that the adaptive batch method offers notable time performance benefits. Despite the extra time needed to compute and adjust the selection probability set during batch selection, this benefit persists.
     
\begin{figure}[!h]
    \centering
    \includegraphics[width=\textwidth]{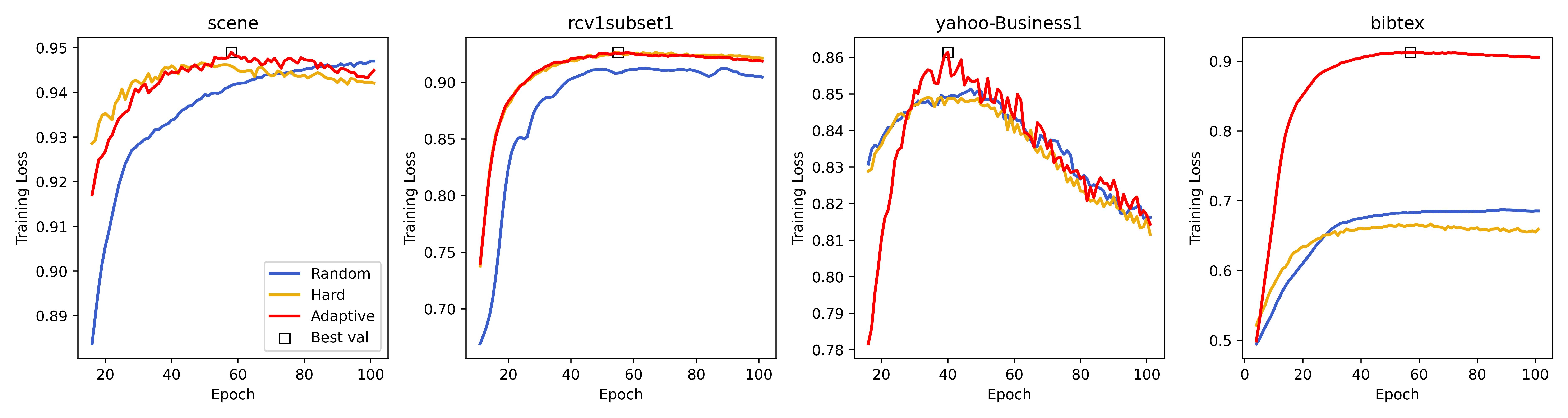}
    \caption{Macro-AUC of validation set for batch selection strategies using DELA with Adam optimizer on different datasets.}
    \label{fig:Macro-AUC}
\end{figure}

     \textbf{Validation Perspective}: As shown in Figure \ref{fig:Macro-AUC}, adaptive batch selection notably enhances model performance. For the scene and rcv1subset1 datasets, all approaches achieve similar Macro-AUC by training's end. Yet, the adaptive batch method attains peak validation epochs more swiftly than random selection strategies. 
    More importantly, the adaptive batch significantly improved the validation set's Macro-AUC on both the yahoo-Business1 and bibtex datasets.

\subsection{Adaptive Batch Selection with Label Correlations}
Table \ref{tab:result1} shows a comparative analysis of adaptive batch selection combining label correlation across different datasets. 
\begin{table}[!ht]
\centering
\resizebox{0.9\textwidth}{!}{
\begin{tabular}{ccc|cc|cc|cc|cc|cc}
\toprule
 & \multicolumn{2}{c}{Macro-F$\uparrow$} & \multicolumn{2}{c}{Micro-F$\uparrow$} & \multicolumn{2}{c}{Macro-AUC$\uparrow$} & \multicolumn{2}{c}{Ranking Loss$\downarrow$} & \multicolumn{2}{c}{Hamming Loss$\downarrow$} & \multicolumn{2}{c}{One Error$\downarrow$} \\ \midrule
 Dataset & Adaptive & AdaC &Adaptive & AdaC & Adaptive & AdaC & Adaptive & AdaC &Adaptive & AdaC & Adaptive & AdaC \\\midrule
 Corel5k & \textbf{0.0963} & 0.0945 & \textbf{0.2081} & 0.2042 & \textbf{0.7621} & 0.7614 & \textbf{0.1532} & 0.1521 & \textbf{0.0223} & 0.0235 & \textbf{0.6520} & 0.6544 \\
 rcv1subset1 & \textbf{0.3184} & 0.3157 & \textbf{0.4544} & 0.4523 & \textbf{0.9194} & 0.9175 & \textbf{0.0571} & 0.0594 & 0.0356 & \textbf{0.0349} & \textbf{0.4305} & 0.4315 \\
 cal500 & 0.0833 & \textbf{0.0940} & 0.3473 & \textbf{0.3491} & 0.5927 & \textbf{0.5962} & 0.2273 & \textbf{0.2234} & 0.1894 & \textbf{0.1852} & 0.1156 & \textbf{0.1135} \\
 yahoo-Business1 & 0.3196 & \textbf{0.3243} & 0.4913 & \textbf{0.4952} & \textbf{0.8062} & 0.8053 & \textbf{0.0343} & 0.0362 & \textbf{0.0167} & 0.0182 & 0.8153 & \textbf{0.8144} \\ \bottomrule
\end{tabular}
}
\caption{The results of adaptive batch selection combining with label correlation.}
\label{tab:result1}
\end{table}
The adaptive batch selection considering label correlations, denoted as "AdaC" performs better in datasets with a high cardinality of labels (cal500), but is ineffective in extremely sparse label spaces (Corel5k). This phenomenon arises from richer mutual information and label associations in high-cardinality settings, boosting the application of combing label correlation.
\subsection{Investigation of Loss \label{bceloss}}
\begin{figure}[!h]
    \centering
    \subfigure[30-th epoch]{\includegraphics[width=0.45\textwidth]{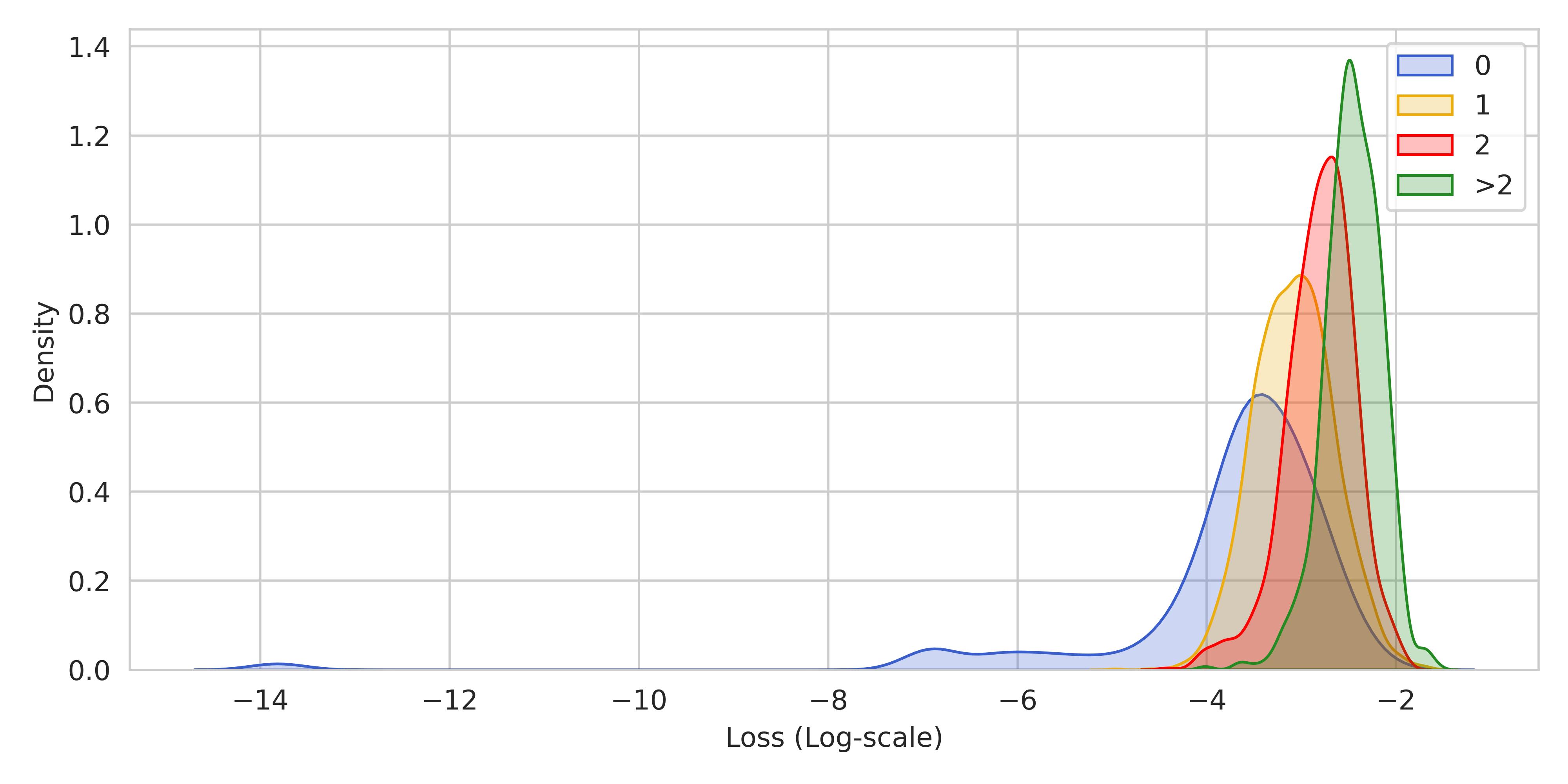}
    }
    \subfigure[70-th epoch]{\includegraphics[width=0.45\textwidth]{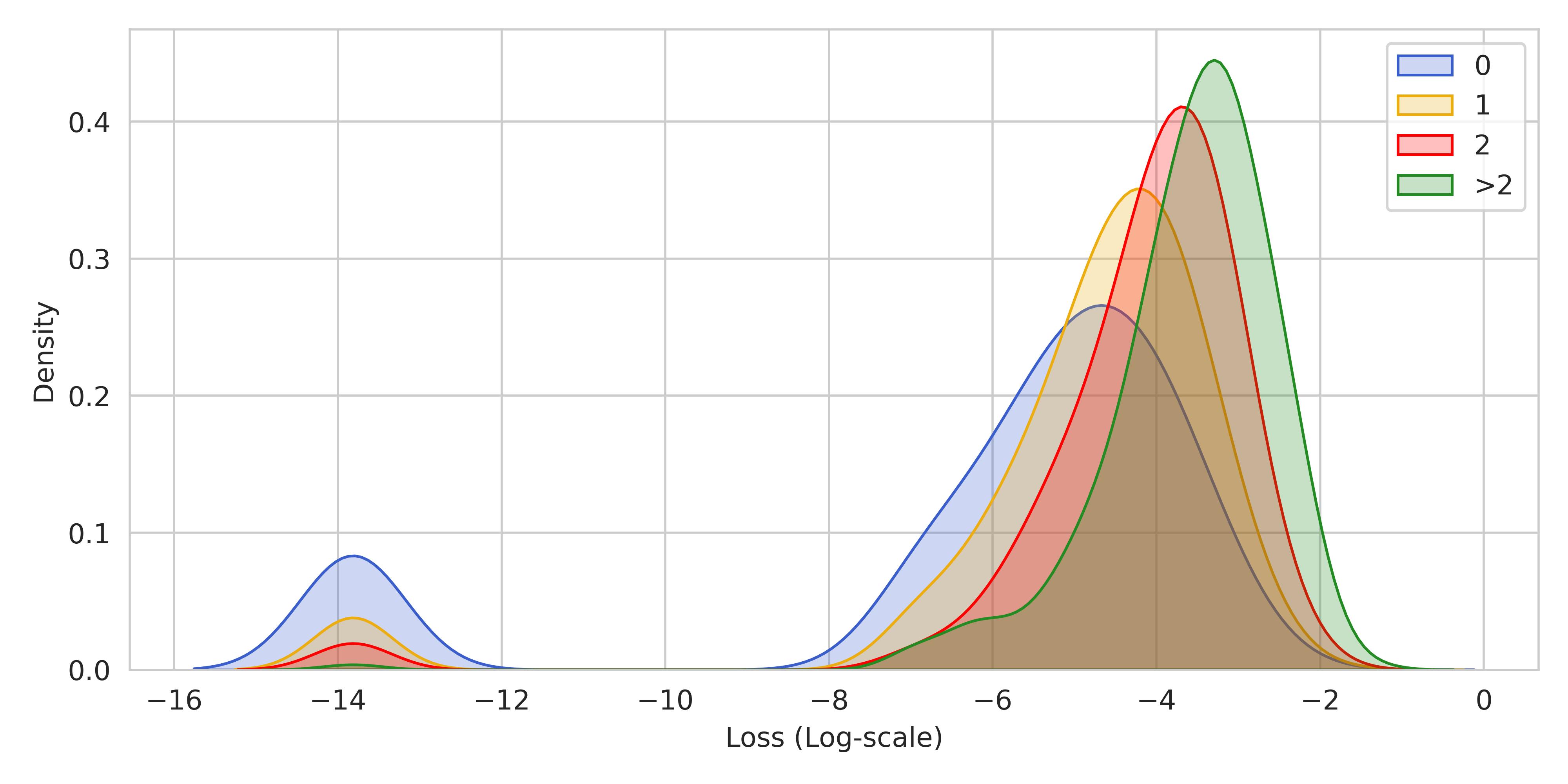}
    } 
    \caption{The loss distribution of minority samples at 30-th epoch and 70-th epoch under adaptive batch selection methods.}
\label{fig:comloss}
\end{figure}

Binary cross-entropy (BCE) loss is a preferred choice for modeling due to its direct measurement of prediction accuracy and has been integrated into the model's loss functions. 
Considering imbalanced losses like Asymmetric Loss \cite{Asymmetric}, introduces significant computational overhead. In contrast, BCE loss offers a more streamlined approach, reducing the need for extra loss calculations and thereby conserving computational resources.
In section \ref{convergencecurve}, while verifying the effectiveness of adaptive batch selection from a global perspective, Figure \ref{fig:comloss} demonstrates from a more fine-grained perspective that the loss of minority label samples under adaptive batch selection can converge better, which provides evidence support for our approach of focusing on hard minority samples.

\begin{figure}[!h]
\centering
    \subfigure[Pearson correlation coefficient]{\includegraphics[width=0.23\textwidth]{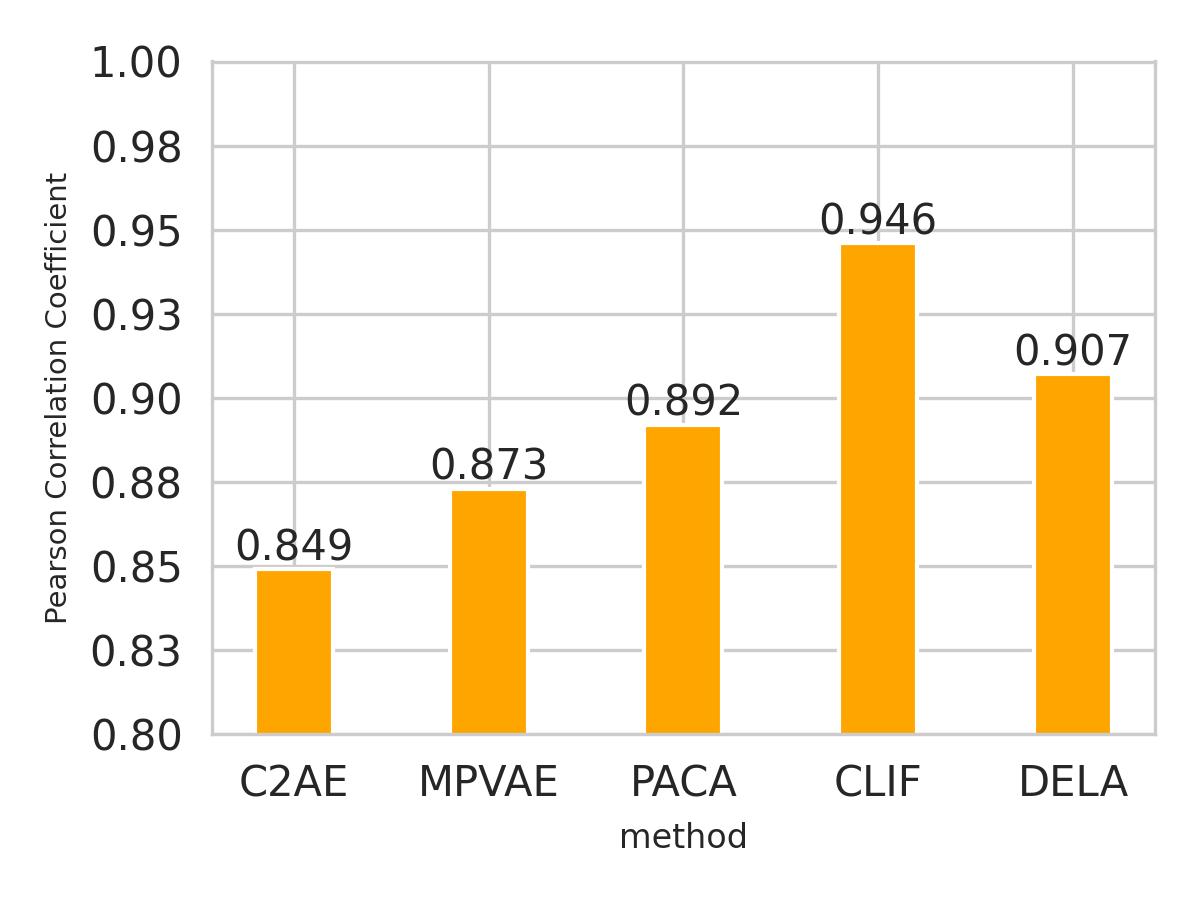}
    \label{fig:consistencea}
    }
    \subfigure[Convergence curves of adaptive batch selection with model loss or BCE loss]{\includegraphics[width=0.72\textwidth]{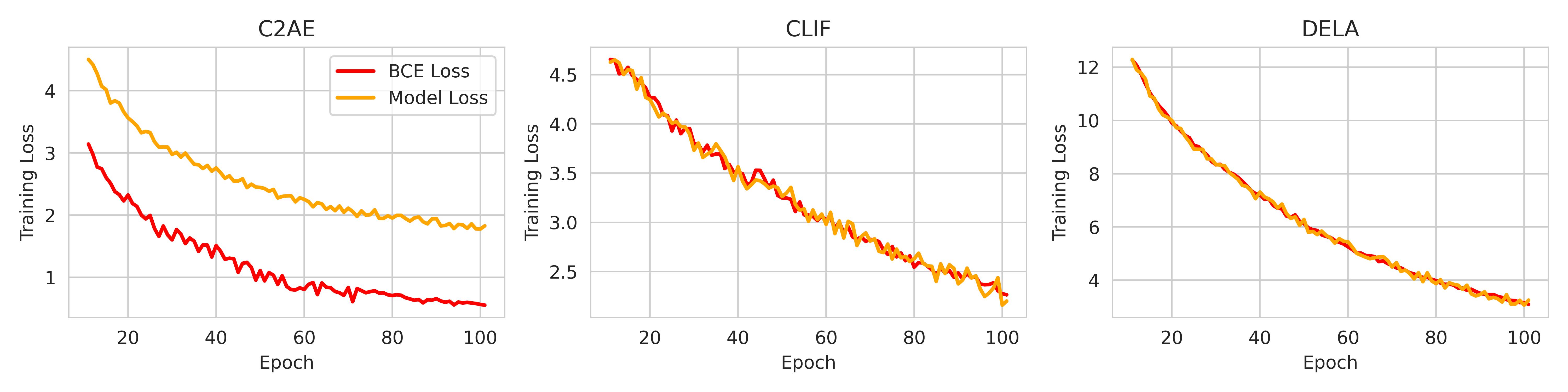}
    \label{fig:consistenceb}
    } 
    \caption{The correlation between adaptive batch selection with the model's loss and BCE loss.}
    \label{fig:consistence}
\end{figure}

Initially, we explore the relationship between the model's loss and BCE loss by analyzing the Pearson correlation coefficient. Figure \ref{fig:consistencea} demonstrates a strong correlation between these two loss metrics. 
Furthermore, Figure \ref{fig:consistenceb} the convergence curve of adaptive batch selection with the model's loss or BCE loss involved in computing $\bm{\ell}$. Both CLIF and DELA models incorporate BCE loss prominently in their loss functions, and giving it more weight than other loss components. This results in similar convergence patterns of the training loss to those seen with adaptive sample selection based on BCE loss. Conversely, C2AE, which does not include BCE loss in its loss function and prioritizes reconstruction loss, shows slower convergence in adaptive batch selection. This slower convergence may be due to the lack of correlation between samples with larger model losses and hard samples under classification tasks.

\subsection{Parameter Analysis}
This section assesses key parameter impacts on our method, particularly selection pressure (\(s_e\)) and batch size variation. Refer to the supplementary appendix for detailed visuals. We conclude that: 1) \(s_e\) may not expedite the convergence of training losses but improves performance metrics by batch setups. 2) the adaptive selection surpasses random selection for batch sizes 256 and 512, proving its effectiveness.

\section{Conclusion}
In this paper, we explore the issue of class imbalance in multi-label classification and its impact on the training of deep models. We propose a novel multi-label batch selection strategy that focuses on class imbalance and hard samples, marking a pioneering effort in this area. Our initial investigations revealed a nuanced relationship between challenging samples and minority labels: samples associated with minority labels tend to incur higher losses during training. By adaptively selecting the Binary Cross Entropy loss with a combination of global and local imbalance weights, we assign greater importance to these challenging samples. Additionally, we refine the ranking strategy through quantization factors to address the issue of smoothness. We also introduce the variant of the adaptive batch selection strategy, such as the chain adaptive batch processing that considers label correlations, which require further and richer experimental validation. Comprehensive experiments demonstrate that our approach leads to better convergence and performance compared to the random batch selection methods used in existing models. Future exploration of batch selection in multi-label learning is intriguing, given the potential "inexplicable" relationships between the model and its mini-batches.


\section*{Ethical Statement}

There are no ethical issues.

%
%
%
\bibliographystyle{splncs04}
\bibliography{ref}
%




\section{Symmetric conditional probability matrix A}
To construct the label relation graph, we denote the graph as \(G = (V, E)\), where \(V\) represents the set of nodes corresponding to the set of class labels, and \(E\) represents the set of edges indicating the co-occurrence relationships between pairs of labels. The adjacency matrix \(\mathbf{A}\) stores each edge's weights, representing the strengths of the co-occurrence relationships. The matrix \(\mathbf{A}\) is defined as the symmetric conditional probability matrix: $A_{ij} = \frac{1}{2} \left[ P(l_j | l_i) + P(l_i | l_j) \right]$, where \(P(l_j | l_i)\) is the probability that label \(l_j\) appears given that label \(l_i\) appears. The diagonal elements of the conditional probability matrix \(P\) are set to 0. We calculate the conditional probability matrix \(P\) on the training set. 
Based on the specified parameter $\left\lceil Card \right\rceil$, we identify the most relevant labels for \(j\) by finding the indices of the top $\left\lceil Card \right\rceil$ highest values in the \(j\)-th row or column of matrix \(\mathbf{A}\).

\section{Appendix}
\subsection{Specific experimental results}

\subsubsection{Selection Pressure}
Selection pressure, denoted as $s_e$, influences the intensity of selection for hard minority samples. When the value of $s_e$ is high, it increases the likelihood of choosing minority samples for the next mini-batch, emphasizing their selection. Conversely, a lower value of $s_e$ reduces the distinction from random batch selection, i.e. making the batch selection more random by diminishing the focus on hard samples.
\begin{figure}[!h]
\centering
\subfigure[yahoo-Arts1]{\includegraphics[width=1\textwidth]{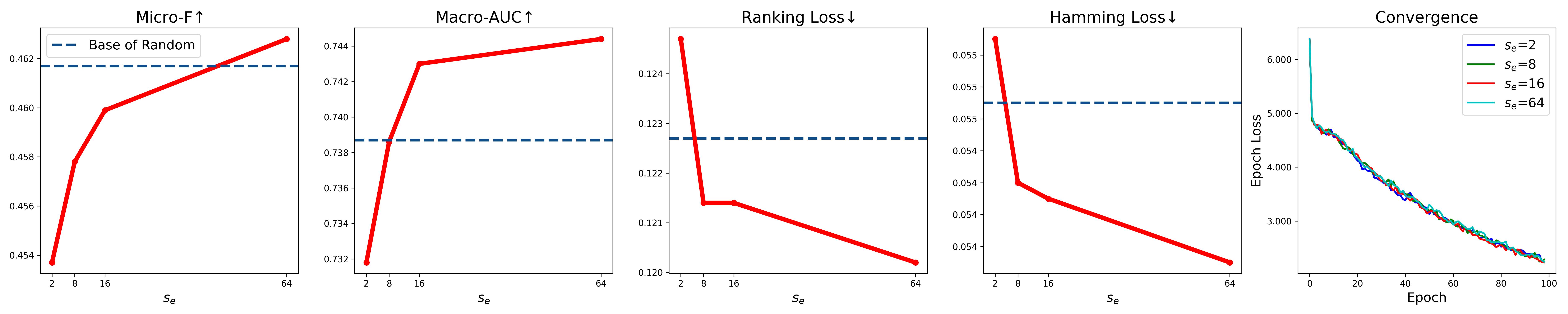}
}
\subfigure[bibtex]{\includegraphics[width=1\textwidth]{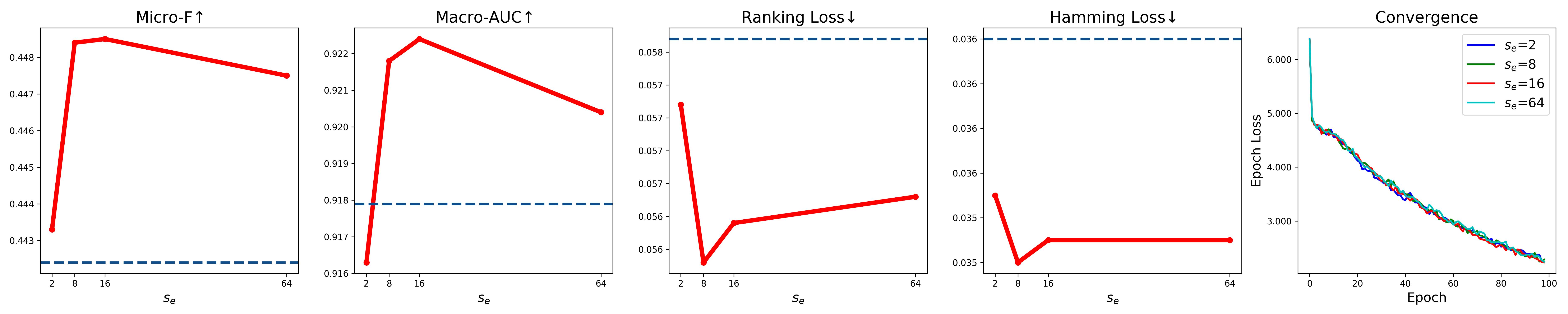}
} 
\caption{The influence of selection pressure $s_e$, where the blue dash line denotes the result of Random method}
\label{fig:se}
\end{figure}

Figure \ref{fig:se} illustrates the impact of varying $s_e$ values on adaptive batch selection across two benchmark datasets in terms of training loss and other evaluation metrics \footnote{Here, we record the result on the first fold}. 
As $s_e$ increases, we observe no significant change in the convergence speed of the training loss. 
This suggests that while altering the $s_e$ might not quicken the decrease of training losses, it can improve evaluation metrics across various batch compositions. 
For the bibtex dataset, a rise in \(s_e\) is directly linked to better evaluation metrics, indicating that focusing on hard minority samples improves model fit. Similarly, for the yahoo-Arts dataset, this positive trend continues as \(s_e\) increases from 2 to 8, but further escalation to 64 diminishes the metrics, hinting at potential overfitting on hard samples.
\subsubsection{Batch size of Model}
To investigate the universality of our adaptive batch selection method, we explore the experimental results of using the DELA model with random batch selection and adaptive batch selection under the condition of batch size=256 and 512. 
\begin{figure}[!h]
\centering
\subfigure[batch size=256]{\includegraphics[width=0.45\textwidth]{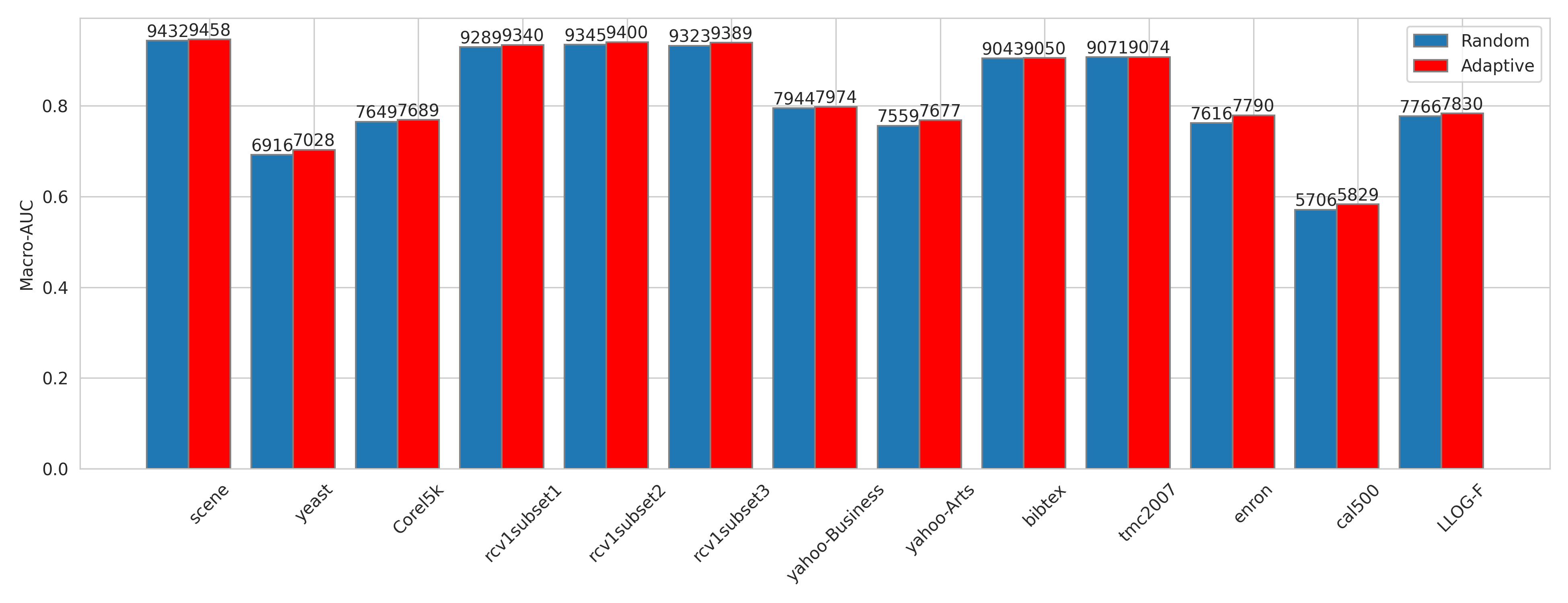}
}
\subfigure[batch size=512]{\includegraphics[width=0.45\textwidth]{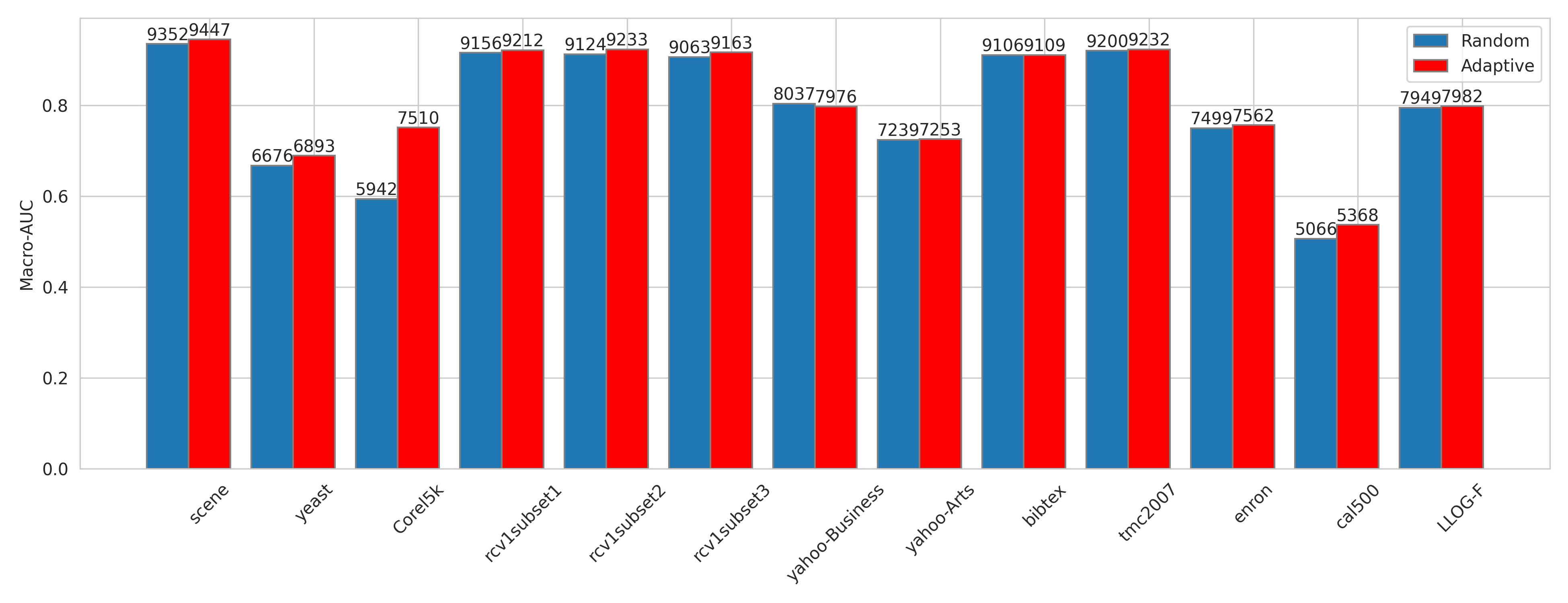}
} 
\caption{The Macro-AUC on different datasets based on DELA with batch size=256 or 512}
\label{fig:batchsize}
\end{figure}
The experimental outcomes demonstrate that our adaptive batch selection method consistently outperforms random batch selection across both batch sizes of 256 and 512, showcasing its superior universality and effectiveness in optimizing model's performance.
\subsection{Proof}
the index $Q(\bm{\ell}_i)$ of each instance is bounded by:
\begin{equation}
0 \leq Q(\bm{\ell}_i) \leq  n
\end{equation}
Then, the lower bound of $p_i \supseteq  P(x_i| D, s_e,q)$ is formulated by
\begin{equation}
0 < \frac{1}{\sum_{i=1}^{n} \exp \left(\log(s_e)/{n} \right)^{Q(x)}} \leq p_i \supseteq P(x_i| D, s_e,q).
\end{equation}
Thus, the sampling distribution of multi-label adaptive selection is strictly positive.
To ensure a gradient estimate is unbiased, the expected value of the gradient estimate under the sampling distribution must equal the true gradient calculated across the entire dataset. Mathematically, this condition can be expressed as:
\begin{equation}
E_{P(x_i | D, s_e, q)}[\nabla L] = \nabla L_{\text{true}},
\end{equation}
where \(\nabla L\) is the gradient of the loss with respect to the model parameters, \(\nabla L_{\text{true}}\) is the true loss computed over the entire dataset, and \(E_{P(x_i | D, s_e, q)}[\cdot]\) denotes the expectation over the sampling distribution. Further, we can assume that if every sample in the training set has a probability of being selected, then the gradient is unbiased. 
Let us consider the gradient estimate $\tilde{G}$ used in combination with the Adam optimizer. Adam, an algorithm for first-order gradient-based optimization of stochastic objective functions, computes adaptive learning rates for each parameter. In its essence, Adam maintains two moving averages for each parameter; one for gradients ($m_t$) and one for the square of gradients ($v_t$) respectively. These moving averages are estimates of the first moment (the mean) and the second moment (the uncentered variance) of the gradients. The unbiasedness of the gradient estimate $\tilde{G}$, when combined with Adam, can be articulated by analyzing the correction step applied to $m_t$ and $v_t$. Specifically, the bias-corrected first and second moment estimates are given by:
\begin{align}
\hat{m}_t &= \frac{m_t}{1 - \beta_1^t}, \\
\hat{v}_t &= \frac{v_t}{1 - \beta_2^t},
\end{align}
where $\beta_1$ and $\beta_2$ are the exponential decay rates for the moment estimates, and $t$ denotes the timestep. The correction factors $(1 - \beta_1^t)$ and $(1 - \beta_2^t)$ counteract the bias towards zero in the initial time steps, ensuring that $\hat{m}_t$ and $\hat{v}_t$ are unbiased estimates of the first and second moments. Therefore, given an unbiased gradient estimate $\tilde{G}$, the application of Adam's bias correction guarantees that the adjusted gradients remain unbiased. 
\end{document}